\documentclass[letterpaper]{article} 
\usepackage{aaai2027} 
\usepackage{amsmath}
\usepackage{amssymb}
\usepackage{amsfonts}
\usepackage[hyphens]{url}  
\usepackage{graphicx} 
\usepackage{natbib}  
\usepackage{caption} 
\usepackage{booktabs}
\usepackage{multirow}

\usepackage{algorithm}
\usepackage{algorithmic}
\usepackage{newfloat}
\usepackage{listings}
\DeclareCaptionStyle{ruled}{labelfont=normalfont,labelsep=colon,strut=off} 
\floatstyle{ruled}
\newfloat{listing}{tb}{lst}{}
\floatname{listing}{Listing}

\nocopyright

\title{DiaRelay: Relaying Dialogue Context with a Constant-Size Memory for Emotion Recognition in Conversation}

\author{
Zihao Zhou\textsuperscript{\rm 1}, Bin Yang\textsuperscript{\rm 1}, Jinghui Qin\textsuperscript{\rm 1}\thanks{Corresponding author}, Kebing Jin\textsuperscript{\rm 2}
}
\affiliations{
    \textsuperscript{\rm 1}Guangdong University of Technology  \\
    \textsuperscript{\rm 2}Guizhou Provincial Laboratory of Big Data, State Key Laboratory of Public Big Data, Guizhou University

    zhouzihao1@mails.gdut.edu.cn, yangbin59@mails.gdut.edu.cn, \\
    qinjinghui@gdut.edu.cn,  kbjin@gzu.edu.cn

}

\begin{document}

\maketitle

\begin{abstract}
Emotion Recognition in Conversation (ERC) requires models to identify subtle emotional cues that are often distributed across distant dialogue turns. Existing methods typically incorporate dialogue history through a fixed context window. However, short windows discard potentially useful long-range evidence, while enlarging the window repeatedly re-encodes overlapping utterances, increases computational and memory costs, and may introduce irrelevant context. Moreover, commonly used parameter-efficient adaptation methods, such as LoRA, mainly introduce fixed low-rank transformations in the feature space and do not explicitly maintain a dialogue-level state or condition their transformations on the evolving conversational context.
To address these limitations, we propose a lightweight adapter, named DiaRelay, to enable large language models (LLMs) to explicitly maintain a dialogue-level memory for accurate ERC. Based on LoRA, DiaRelay introduces two extra tightly collaborative components, Selective Relay Memory Transition and Dual-axis Relay Memory Read. Selective Relay Memory Transition progressively aggregates useful historical evidence into a bounded relay memory and propagates it across successive utterance predictions. This allows earlier emotional cues to influence later predictions after they leave the local context window, without re-encoding the complete dialogue history or expanding the backbone context length. Dual-axis Relay Memory Read uses the propagated memory to dynamically modulate low-rank feature transformations, enabling context-dependent representation adaptation without test-time gradient updates. Extensive experiments show that DiaRelay can achieve state-of-the-art weighted F1 and accuracy on MELD while obtaining competitive results on IEMOCAP with only an extra $\sim$7.1M trainable parameters, indicating the effectiveness and generalizability of our DiaRelay in enhancing LLM-based emotional understanding. 


\end{abstract}


\section{Introduction}

Emotion Recognition in Conversation (ERC) aims to identify the emotion expressed by each utterance in a dialogue. It plays an important role in affective computing and supports a wide range of applications, including conversational agents, social media analysis, and human--computer interaction. Unlike isolated emotion classification, ERC requires models to interpret each utterance in relation to its conversational context. A short or semantically ambiguous utterance may convey different emotions depending on preceding events, speaker interactions, and previously expressed attitudes. Therefore, effectively modeling dialogue history is a central challenge in ERC \cite{erc_survey,dialoguernn,dialoguegcn}.

Existing ERC methods capture contextual information using recurrent networks, graph-based reasoning, attention mechanisms, and pretrained language models \cite{dialoguernn,dialoguegcn,dialogxl,instructerc}. With the increasing adoption of large language models, a common strategy is to concatenate the current utterance with a fixed number of preceding utterances as context and deploy parameter-efficient fine-tuning (PEFT) technology to endow LLMs with the capability to predict the emotion inherent in the current utterance. 
However, this strategy has the following limitation. A short context window may exclude potentially useful historical evidence once earlier utterances fall outside the context window.
Enlarging the context window provides broader historical coverage, but 
it increases computational and memory costs, and may introduce irrelevant context. Although parameter-efficient adaptation methods such as LoRA \cite{lora} are deployed to enable LLMs to understand emotions and reason efficiently, this limitation can not be resolved by these methods since the learned mappings of LoRA or its variants are shared across dialogue turns and do not explicitly maintain a persistent dialogue-level memory that is crucial for accurate emotion recognition in conversation with limited context.  
Consequently, once an utterance leaves the explicit context window, standard PEFT technology can not preserve useful cues antecedent to the current context window or condition later transformations on the evolving dialogue history.

This limitation motivates us to develop a new memory-augmented adaptation mechanism that complements explicit local context modeling with a compact dialogue-level memory. Such a mechanism should preserve useful historical information beyond the local context window, process each utterance adaptively according to the accumulated conversation, and maintain a bounded memory size as the dialogue grows. Meanwhile, it also should retain the parameter efficiency of low-rank adaptation and avoid gradient-based parameter updates during inference.

To this end, we propose \textbf{DiaRelay}, a lightweight memory-augmented adapter that enables LLMs to explicitly maintain and exploit dialogue-level memory for ERC. Based on LoRA, DiaRelay complements a fixed local context window with a bounded cross-utterance constant-size relay memory, which is progressively propagated across successive utterance predictions. Rather than retaining an ever-growing sequence of historical representations, DiaRelay compresses conversation history into a constant-size memory and uses it to condition the low-rank adaptation applied to the current utterance. DiaRelay contains two tightly collaborative components: \textbf{Selective Relay Memory Transition (SRMT)} and \textbf{Dual-axis Relay Memory Read (DaRMR)}. SRMT progressively updates the relay memory through a gated error-corrective operation. It selectively incorporates useful information from the current utterance while retaining useful historical content, allowing the memory to evolve throughout the conversation without increasing the memory size. DaRMR retrieves complementary information from the propagated memory and produces context-dependent low-rank corrections for the query and output pathways of self-attention. In this way, the backbone representation is dynamically adapted according to the accumulated dialogue context rather than relying solely on a fixed low-rank transformation.
The two components operate under a "\textbf{read-before-write}" paradigm. When predicting the $t$-th utterance, DaRMR accesses only the relay memory of the previous $t-1$ turns. After the prediction, SRMT incorporates the current utterance into the relay memory for memory updating. This paradigm prevents the current utterance from being written into the memory before its own prediction, while allowing it to provide historical information for subsequent utterances. As a result, earlier emotional cues can continue to affect later predictions after leaving the explicit context window, without expanding the context length of inputs, re-encoding the complete dialogue history, or performing test-time gradient updates.
Extensive experiments on MELD and IEMOCAP under a unified history-only setting show that DiaRelay improves both accuracy and weighted F1 over strong LoRA-based fixed-window baselines, indicating the effectiveness and generalizability of our DiaRelay in enhancing LLM-based emotional understanding. In particular, DiaRelay achieves state-of-the-art weighted F1 on MELD while obtaining competitive performance on IEMOCAP.  


Overall, our main contributions are threefold:
\begin{itemize}
    \item We propose \textbf{DiaRelay}, a lightweight memory-augmented adapter that complements a fixed local context window with a bounded cross-utterance constant-size relay memory. DiaRelay enables historical information to persist across successive utterance predictions without expanding the input context length or requiring test-time parameter updates.
    
    \item In DiaRelay, we introduce \textbf{SRMT} and \textbf{DaRMR}. The SRMT selectively updates the relay memory through a gated error-corrective transition, while the DaRMR retrieves complementary memory information to generate context-dependent low-rank corrections for the query and output pathways. Together, they establish a "read-before-write" paradigm for dialogue-level memory propagation and utilization.
    
    \item Extensive experiments on MELD and IEMOCAP demonstrate that DiaRelay outperforms strong LoRA-based fixed-window baselines in both accuracy and weighted F1, indicating the effectiveness and generalizability of our DiaRelay in enhancing LLM-based emotional understanding. DiaRelay achieves state-of-the-art weighted F1 on MELD while adding only $\sim$7.1M trainable parameters, approximately $0.09\%$ of the Qwen3-8B backbone.
\end{itemize}

\section{Related Work}

\noindent\textbf{Contextual Modeling for ERC.}
Modeling conversational context has long been a central topic in Emotion Recognition in Conversation (ERC). Early approaches mainly rely on recurrent architectures to propagate contextual and speaker-specific representations along a dialogue. For example, DialogueRNN \cite{dialoguernn} tracks the evolving states of individual speakers, while DialogueCRN \cite{dialoguecrn} performs iterative retrieval and reasoning to integrate emotional clues from conversational history. Graph-based methods further represent utterances as nodes and model conversational dependencies through explicit message passing. DialogueGCN \cite{dialoguegcn} captures intra- and inter-speaker relations using a conversation graph, whereas DAG-ERC \cite{dagerc} combines graph propagation with recurrent information flow to connect nearby context with long-distance dialogue background. Although these methods effectively model contextual dependencies, they rely on task-specific recurrent reasoning or explicitly constructed utterance-level structures. Such designs are not directly tailored to parameter-efficient adaptation of large language models, where the dialogue history is commonly re-encoded as part of the input for each target utterance.

Several studies have introduced memory-related mechanisms into ERC. DialogXL~\cite{dialogxl} modifies the recurrence mechanism of XLNet from the segment level to the utterance level and replaces its original self-attention with dialog-aware self-attention, allowing longer historical representations to be retained across utterances. However, this design is closely coupled with the internal recurrence and attention architecture of XLNet, making it difficult to transfer directly to a general LoRA-adapted LLM without modifying the backbone. CoMPM \cite{compm} combines a context model with an additional pretrained memory extractor to obtain speaker-specific information from previous utterances. Its memory primarily serves as an additional pretrained representation and requires a separate memory extraction pathway alongside the context encoder. In contrast, DiaRelay maintains an online and bounded relay memory that evolves after each utterance prediction. It neither constructs an explicit conversation graph nor replaces the backbone attention mechanism, and directly uses the propagated memory to condition low-rank feature transformations.

\noindent\textbf{LLM-based ERC.} 
Recent studies have explored large language models for ERC by reformulating emotion classification as an instruction-following or generative task. InstructERC \cite{instructerc} introduces a retrieval-based template and auxiliary speaker identification and emotion prediction objectives to incorporate multi-granularity dialogue supervision. Subsequent methods further enrich LLMs with additional speaker knowledge and reasoning supervision. LaERC-S \cite{laercs} prompts LLMs to derive speaker characteristics, such as mental states and behaviors, and adopts a two-stage learning procedure to inject these characteristics into emotion prediction. CoE \cite{coe} progressively integrates conversational clues through role-playing, speaker identification, and emotion reasoning tasks under a multi-stage auxiliary learning strategy. PRC-Emo \cite{prcemo} combines emotion-sensitive prompting, demonstration retrieval, and curriculum learning, supported by a dedicated repository containing retrieved and generated dialogue demonstrations.

These methods substantially improve the ability of LLMs to interpret emotional cues, but their gains mainly arise from richer input construction, additional knowledge, auxiliary supervision, or multi-stage training. Dialogue history is still primarily conveyed through the explicit context constructed for each target utterance, requiring overlapping historical content to be repeatedly encoded across successive predictions. Moreover, retrieval-based demonstrations, generated speaker descriptions, and auxiliary reasoning objectives introduce additional data construction or training complexity. They do not explicitly maintain a compact state that continuously evolves across utterance predictions.

Parameter-efficient fine-tuning methods such as LoRA \cite{lora} make the adaptation of LLMs substantially more affordable by introducing trainable low-rank transformations while freezing most backbone parameters. Nevertheless, the learned low-rank mappings are shared across dialogue turns and provide no explicit mechanism for carrying accumulated dialogue information from one prediction to the next. DiaRelay addresses this orthogonal limitation by propagating a bounded relay memory across utterances and using it to condition low-rank transformations. It therefore complements explicit local context modeling without relying on retrieved demonstrations, externally generated speaker profiles, additional reasoning labels, or test-time gradient updates.

\begin{figure*}[t]
    \centering
    \includegraphics[width=0.98\textwidth]{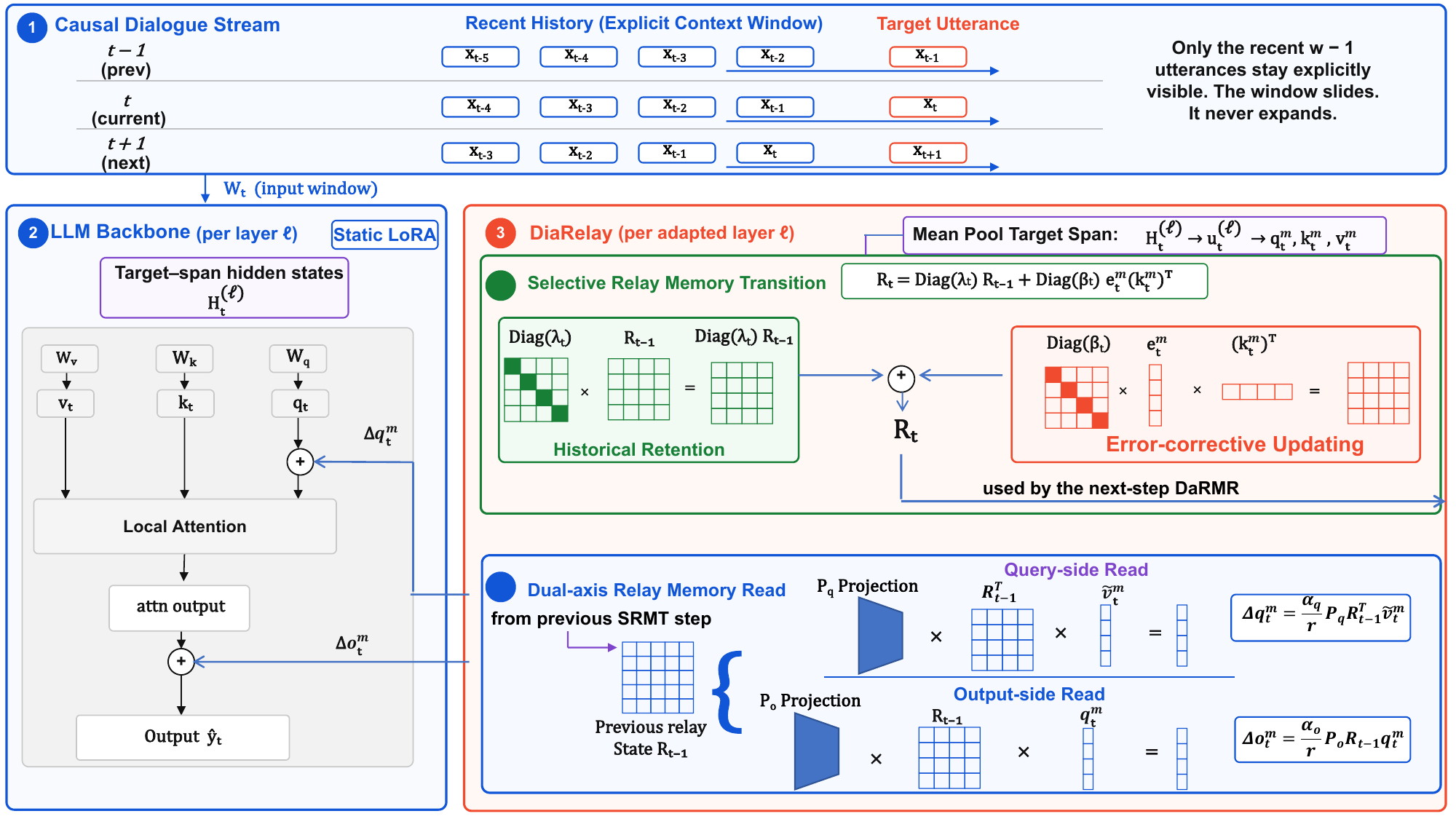}
    \caption{Overall architecture of DiaRelay.}
    \label{fig:diarelay_architecture}
\end{figure*}

\section{DiaRelay}
\label{sec:method}
\subsection{Problem Formulation \& Framework Overview}
\label{sec:method_overview}

Given a dialogue containing $T$ utterances, we denote it as $\mathcal{D}
=
\left\{
(x_t,s_t)
\right\}_{t=1}^{T},$
where $x_t$ and $s_t$ denote the text and speaker of the $t$-th utterance, respectively. The goal of ERC is to predict an emotion label
$y_t\in\mathcal{Y}$ for each utterance $x_t$ according to dialogue context $\mathcal{C}_t$, where $\mathcal{Y}$ are the emotion label space.
For the current utterance $x_t$, the dialogue context $\mathcal{C}_t$, which is also the model input, is constructed from continuous utterances in a fixed contextual window: \[
\mathcal{C}_t
=
\left[
(x_{\max(1,t-w+1)},s_{\max(1,t-w+1)}),
\ldots,
(x_t,s_t)
\right],
\]
where $w$ is the contextual window size. 

Based on LoRA, DiaRelay equips each adapted Transformer layer with an independent relay memory. Let $\mathcal{L}_{\mathrm{D}}$ denote the set of layers augmented with DiaRelay. At the $\ell$-th adapted layer, the relay memory is represented as 
$\mathbf{R}_{t}^{(\ell)}
\in
\mathbb{R}^{r\times r}$,
where $r$ is the relay rank. The memory is initialized at the beginning of each dialogue as $\mathbf{R}_{0}^{(\ell)}
=
\mathbf{0}$. 

After tokenizing $\mathcal{C}_t$, let $\mathbf{H}_{t}^{(\ell)}
\in
\mathbb{R}^{N_t\times d}$
denote the hidden states entering the $\ell$-th self-attention layer, where $N_t$ is the number of input tokens and $d$ is the output hidden dimension. Let $\mathcal{I}_t$ denote the token indices corresponding to the target utterance $x_t$. Its layer-wise representation $\mathbf{u}_{t}^{(\ell)} \in
\mathbb{R}^{d}$ is obtained by mean pooling over the target span as follows:
\begin{equation}
\mathbf{u}_{t}^{(\ell)}
=
\frac{1}{|\mathcal{I}_t|}
\sum_{i\in\mathcal{I}_t}
\mathbf{H}_{t,i}^{(\ell)}.
\label{eq:utterance_representation}
\end{equation}

As shown in Figure~\ref{fig:diarelay_architecture}, DiaRelay contains two tightly collaborative components. \textbf{Selective Relay Memory Transition (SRMT)} converts the target-utterance representation into low-dimensional memory vectors and updates the bounded relay memory. \textbf{Dual-axis Relay Memory Read (DaRMR)} retrieves the information accumulated in the relay memory and maps it to query-side and output-side corrections for self-attention calibration. At step $t$, $\mathbf{R}_{t-1}$ denotes the historical memory available before the current utterance/update, while $\mathbf{R}_{t}$ denotes the updated memory after incorporating the current utterance representation.

Since the formulations for different layers are the same, we omit the layer superscript $(\ell)$ in the following derivations for simplicity when no ambiguity arises.

\subsection{Selective Relay Memory Transition (SRMT)}
\label{sec:selective_transition}

SRMT converts the target-utterance representation into a compact relay space and progressively integrates the resulting information into a bounded constant-size relay memory. SRMT first constructs low-dimensional coordinates for memory writing and reading via \emph{Relay-Space Projection}. Then, SRMT applies \emph{Selective Memory Relay} to update the memory through dimension-wise selection and error-corrective information propagation.

\noindent\textbf{Relay-Space Projection.}
Since the layer-wise utterance representation
$\mathbf{u}_t\in\mathbb{R}^{d}$ lies in the high-dimensional feature space and cannot be directly incorporated into the compact relay memory, we project it into 3 low-dimensional relay vectors $\mathbf{q}_{t}^{m} \in
\mathbb{R}^{r}$, $\mathbf{k}_{t}^{m} \in
\mathbb{R}^{r}$, and $\mathbf{v}_{t}^{m} \in
\mathbb{R}^{r}$:
\begin{align}
\mathbf{q}_{t}^{m}
&=
\operatorname{Norm}
\left(
\tanh
\left(
\mathbf{W}_{q}^{m}\mathbf{u}_{t}
\right)
\right),
\label{eq:relay_query}
\\
\mathbf{k}_{t}^{m}
&=
\operatorname{Norm}
\left(
\tanh
\left(
\mathbf{W}_{k}^{m}\mathbf{u}_{t}
\right)
\right),
\label{eq:relay_key}
\\
\mathbf{v}_{t}^{m}
&=
\mathbf{W}_{v}^{m}\mathbf{u}_{t},
\label{eq:relay_value}
\end{align}
where $\mathbf{W}_{q}^{m}$, $\mathbf{W}_{k}^{m}$, and $\mathbf{W}_{v}^{m}$ are learnable projection matrices with the size of $r\times d$
, $\operatorname{Norm}(\cdot)$ denotes $\ell_2$ normalization, and $tanh$ is Tanh activation function. Here, $\mathbf{k}_{t}^{m}$ and $\mathbf{v}_{t}^{m}$ form the
key--value association written into the relay memory, while
$\mathbf{q}_{t}^{m}$ serves as the read vector used by the
subsequent DaRMR component.

\noindent\textbf{Selective Memory Relay.}
Given the previous relay memory
$\mathbf{R}_{t-1}\in\mathbb{R}^{r\times r}$ and the current relay
vectors, the memory is updated as follows:
\begin{equation}
\mathbf{R}_{t}
=
\operatorname{Diag}
\left(
\boldsymbol{\lambda}_{t}
\right)
\mathbf{R}_{t-1}
+
\operatorname{Diag}
\left(
\boldsymbol{\beta}_{t}
\right)
\left(
\mathbf{e}_{t}^{m}
\right)
{\mathbf{k}_{t}^{m}}^{\top},
\label{eq:memory_transition}
\end{equation}
where $\boldsymbol{\beta}_{t}$ is the update gate and $\boldsymbol{\lambda}_{t}$ is its complementary retention gate. $\mathbf{e}_{t}^{m}$ is the relay operation. $\boldsymbol{\beta}_{t}$ is generated from the current utterance representation as follows:
\begin{equation}
\boldsymbol{\beta}_{t}
=
\sigma
\left(
\mathbf{W}_{\beta}\mathbf{u}_{t}
+
\mathbf{b}_{\beta}
\right)
\in
(0,1)^{r},
\label{eq:write_gate}
\end{equation}
and its complementary retention gate is defined as
\begin{equation}
\boldsymbol{\lambda}_{t}
=
\mathbf{1}
-
\boldsymbol{\beta}_{t},
\label{eq:retention_gate}
\end{equation}
where
$\mathbf{W}_{\beta}\in\mathbb{R}^{r\times d}$ and
$\mathbf{b}_{\beta}\in\mathbb{R}^{r}$.
These two gates operate independently over the relay value dimensions.
Accordingly, $\boldsymbol{\lambda}_{t}$ controls how much of the historical memory
is retained, whereas $\boldsymbol{\beta}_{t}$ controls how strongly
the current information modifies each memory dimension. This
dimension-wise gating constitutes the \emph{selective} operation of SRMT. 
Inspired by the delta rule \cite{schlag2021linear}, we construct an
 error-corrective write signal:
\begin{equation}
\mathbf{e}_t^m
=
\mathbf{v}_t^m
-
\mathbf{R}_{t-1}\mathbf{k}_t^m.
\end{equation}
Here,
$\mathbf{R}_{t-1}\mathbf{k}_{t}^{m}$ is the value that the existing
memory associates with the current key direction, and
$\mathbf{e}_{t}^{m}$ is the component of the current value that is not
recovered from the previous memory.  This operation allows SRMT to relay this residual component through 
$\mathbf{e}_{t}^{m}{\mathbf{k}_{t}^{m}}^{\top}$ in Equation~\eqref{eq:memory_transition}, rather than
indiscriminately accumulating the complete current value.

Thus, each relay dimension independently balances historical
retention and residual write-in, allowing the bounded relay memory to preserve
stable dialogue information while continuously incorporating newly
observed content.

\subsection{Dual-axis Relay Memory Read (DaRMR)}
\label{sec:dual_axis_read}
DaRMR retrieves the information accumulated in
the relay memory and converts it into two complementary corrections for
self-attention. Specifically, the memory
$\mathbf{R}_{t-1}\in\mathbb{R}^{r\times r}$ is read along two
directions to generate a query-side correction
$\Delta\mathbf{q}_t^{m}$ and an output-side correction
$\Delta\mathbf{o}_t^{m}$.

\noindent\textbf{Dual-axis Memory Readout.}
The two memory-conditioned corrections $\Delta\mathbf{q}_t^{m}$ and $\Delta\mathbf{o}_t^{m}$ are computed as follows:
\begin{align}
\Delta\mathbf{q}_{t}^{m}
&=
\frac{\alpha_q}{r}
\mathbf{P}_{q}
\mathbf{R}_{t-1}^{\top}
\operatorname{Norm}
\left(
\tanh
\left(
\mathbf{v}_{t}^{m}
\right)
\right),
\label{eq:query_memory_correction}
\\
\Delta\mathbf{o}_{t}^{m}
&=
\frac{\alpha_o}{r}
\mathbf{P}_{o}
\mathbf{R}_{t-1}
\mathbf{q}_{t}^{m}.
\label{eq:output_memory_correction}
\end{align}
where
$\mathbf{P}_{q}\in\mathbb{R}^{d_q\times r}$ and
$\mathbf{P}_{o}\in\mathbb{R}^{d\times r}$ are trainable projections,
while $\alpha_q$ and $\alpha_o$ control the correction scales. 
Equation~\eqref{eq:query_memory_correction} reads the relay memory along its
value-to-key direction and produces a correction for the attention
query. Equation~\eqref{eq:output_memory_correction} reads the memory along its
key-to-value direction and retrieves a correction for the attention
output.

\noindent\textbf{Memory-conditioned Attention Correction.}
The two memory readouts $\Delta\mathbf{q}_t^{m}$ and $\Delta\mathbf{o}_t^{m}$ are incorporated into the query and output pathways of self-attention for attention correction. For each token position
$i\in\mathcal{I}_t$ belonging to the target utterance, the final query
representation is defined as follows:
\begin{equation}
\widetilde{\mathbf{q}}_{t,i}
=
\mathbf{q}_{t,i}^{0}
+
\Delta\mathbf{q}_{t,i}^{\mathrm{L}}
+
\Delta\mathbf{q}_{t}^{m},
\label{eq:memory_corrected_query}
\end{equation}
where $\mathbf{q}_{t,i}^{0}$ is the original backbone query,
$\Delta\mathbf{q}_{t,i}^{\mathrm{L}}$ is the LoRA query residual, and
$\Delta\mathbf{q}_{t}^{\mathrm{m}}$ is the query-side correction
retrieved from the relay memory.

By using the corrected query in the standard attention computation, the
final output representation is defined as follows:
\begin{equation}
\widetilde{\mathbf{o}}_{t,i}
=
\mathbf{o}_{t,i}^{0}
+
\Delta\mathbf{o}_{t,i}^{\mathrm{L}}
+
\Delta\mathbf{o}_{t}^{m},
\label{eq:memory_corrected_output}
\end{equation}
where $\mathbf{o}_{t,i}^{0}$ and
$\Delta\mathbf{o}_{t,i}^{\mathrm{L}}$ denote the original backbone
output and its LoRA residual, respectively, while
$\Delta\mathbf{o}_{t}^{\mathrm{m}}$ is the output-side correction
retrieved from the relay memory.

The query-side correction steers the attention computation according
to the propagated dialogue memory, while the output-side correction
injects the retrieved historical information into the resulting
representation. The corrections are applied only to the target
utterance span, leaving the representations of the explicit historical
context unchanged.

\subsection{Learning Objective}
\label{sec:objective_complexity}

DiaRelay preserves the original generative objective of LLM-based ERC. Let
$\mathbf{y}_t=(y_{t,1},\ldots,y_{t,M_t})$
denote the token sequence representing the ground-truth emotion label of utterance $x_t$. $M_t$ is the length of the target emotion label after tokenization.
The training objective is the autoregressive negative log-likelihood:
\begin{equation}
\mathcal{L}_{\mathrm{ERC}}
=
-
\sum_{t=1}^{T}
\sum_{j=1}^{M_t}
\log
p_{\theta}
\left(
y_{t,j}
\mid
y_{t,<j},
\mathcal{C}_t,
\left\{
\mathbf{R}_{t-1}^{(\ell)}
\right\}_{\ell\in\mathcal{L}_{\mathrm{D}}}
\right).
\label{eq:training_objective}
\end{equation}
It is noted that in our DiaRelay, no additional memory supervision, retrieved demonstration, speaker-profile label, or reasoning annotation is required. During both training and inference, the relay memory is propagated sequentially within each dialogue and reset at dialogue boundaries.

\section{Experiments}

\subsection{Experimental Settings}

\noindent\textbf{Datasets.} We use two ERC datasets: IEMOCAP \cite{busso2008iemocap},
which consists of dyadic conversations, and MELD
\cite{poria2019meld}, a multiparty conversation dataset derived
from the TV series \textit{Friends}. Dataset statistics are shown
in Table~\ref{tab:dataset_stats}.

\begin{table}[t]
\centering
\small
\setlength{\tabcolsep}{8pt}
\renewcommand{\arraystretch}{1.08}
\begin{tabular}{@{}llrr@{}}
\toprule
Dataset & Partition & Utterances & Dialogues \\
\midrule
\multirow{2}{*}{IEMOCAP}
& train + valid & 5,810  & 120 \\
& test          & 1,623  & 31  \\
\midrule
\multirow{2}{*}{MELD}
& train + valid & 11,098 & 1,152 \\
& test          & 2,610  & 280 \\
\bottomrule
\end{tabular}
\caption{Statistics of the two datasets.}
\label{tab:dataset_stats}
\end{table}

\begin{table}[t]
\centering
\Huge
\resizebox{\linewidth}{!}{
\begin{tabular}{@{}lllcccc@{}}
\toprule
\multirow{2}{*}{Method}
& \multirow{2}{*}{Backbone}
& \multirow{2}{*}{Mod}
& \multicolumn{2}{c}{IEMOCAP}
& \multicolumn{2}{c}{MELD} \\
\cmidrule(lr){4-5}
\cmidrule(lr){6-7}
& & & Acc. & W-F1 & Acc. & W-F1 \\
\midrule

DialogueRNN\textsuperscript{$\dagger$}
& CNN
& T
& 63.40
& 62.75
& 59.54
& 57.03 \\

DialogueGCN\textsuperscript{$\dagger$}
& CNN
& T
& 65.25
& 64.18
& 59.46
& 58.10 \\

COSMIC\textsuperscript{$\dagger\ddagger$}
& RoBERTa
& T
& --
& 65.28
& --
& 65.21 \\

MMLA (SFT)
& Llama-3.2-3B
& T
& 50.00
& 49.24
& 64.41
& 63.40 \\

MSE-Adapter\textsuperscript{$\dagger$}
& ChatGLM3-6B 
& TAV
& --
& --
& 66.23
& 65.13 \\

InstructERC\textsuperscript{$\ddagger$}
& LLaMA2-7B
& T
& --
& 71.39
& --
& 69.15 \\

BiosERC-7B\textsuperscript{$\dagger\ddagger$}
& LLaMA2-7B
& T
& --
& 68.72
& --
& 69.02 \\

MSG-LLM\textsuperscript{$\ddagger$}
& LLaMA2-7B
& T
& --
& 72.02
& --
& 69.14 \\

LaERC-S\textsuperscript{$\dagger\ddagger$}
& LLaMA2-7B
& T
& --
& 72.40
& --
& 69.27 \\

SpeechCueLLM
& LLaMA2-7B
& TA
& --
& \textbf{72.60}
& --
& 67.60 \\

Causal-ERC (T)
& LLaMA-3.1-8B 
& T
& 69.19
& 69.13
& 69.43
& 68.10 \\

PRC-Emo (Causal)\textsuperscript{$\ddagger,*$}
& Qwen2.5-7B/Qwen3-8B
& T
& 69.75 & 69.76
& 70.61 & 69.63 \\

\midrule

\textbf{DiaRelay (Ours)}
& Qwen3-4B
& T
& \underline{67.34}
& \underline{67.08}
& \underline{67.56}
& \underline{65.53} \\

\textbf{DiaRelay (Ours)}
& Qwen3-8B
& T
& \textbf{69.93}
& 70.01
& \textbf{71.15}
& \textbf{70.06} \\

\bottomrule
\end{tabular}}
\caption{Comparison with representative ERC methods on IEMOCAP and MELD. \textbf{Bold} values indicate the best overall results, while
\underline{underlined} values indicate the best results among methods
using backbones no larger than 6B.
\textsuperscript{$\dagger$} indicates the use of future or full-dialogue
information. \textsuperscript{$\ddagger$} indicates the use of additional retrieved,
generated, commonsense, biography, or speaker-related knowledge. $*$ denotes our causal reimplementation of PRC-Emo, where future utterances are excluded. Mod denotes modality, T denotes text modality, A denotes audio modality, and V denotes visual modality. For PRC-Emo (Causal), the two backbones correspond to
IEMOCAP and MELD, respectively.
} 
\label{tab:main_results}


\end{table}

\noindent\textbf{Metrics.} We use weighted F1 (W-F1) as the primary evaluation metric
and additionally report accuracy (Acc) for performance comparison. Since Macro-F1 is not consistently reported
by existing baselines, we include it only in the ablation studies to provide an additional class-balanced comparison among our model variants.

\noindent\textbf{Baselines.}
We compare DiaRelay with representative ERC methods,
including conventional neural models such as
DialogueRNN~\cite{dialoguernn} and
DialogueGCN~\cite{dialoguegcn}, pretrained language-model
methods such as COSMIC~\cite{cosmic}, and recent LLM-based
methods such as MMLA~\cite{mmla},
InstructERC~\cite{instructerc}, BiosERC~\cite{bioserc},
MSG-LLM~\cite{ding-etal-2025-msg}, LaERC-S~\cite{laercs},
SpeechCueLLM~\cite{speechcuellm}, 
Causal-ERC~\cite{causalerc}, and PRC-Emo~\cite{prcemo}. For a fair comparison under the history-only setting, we further
reimplement PRC-Emo in a causal manner by excluding future
utterances from its dialogue context. We also include
MSE-Adapter~\cite{mseadapter} as a multimodal comparison
method.
As these methods differ in their available contextual information and auxiliary resources,
we explicitly mark those methods using
future or full-dialogue information, external knowledge, or
additional audio and visual modalities in the main experimental Table~\ref{tab:main_results}.
The brief introduction to baselines is provided in \textbf{Supplementary Material A1}. 

\noindent\textbf{Implementation Details.}
We implement DiaRelay with Qwen3-4B and Qwen3-8B \cite{qwen3}, where Qwen3-8B serves as the primary backbone and Qwen3-4B is used to evaluate its effectiveness with a more compact language model. LoRA is applied for parameter-efficient adaptation. The explicit input context contains at most four preceding utterances and the current target utterance. DiaRelay uses a relay rank of 8 and is inserted into all Transformer layers. The relay state is propagated along each dialogue and reset at dialogue boundaries.

All experiments are conducted on a single NVIDIA 3090 GPU. The main results are averaged over three runs with random seeds. Detailed optimization and training configurations are
provided in \textbf{Supplementary Material A2}. 

\begin{table*}[t]
\centering
\small
\setlength{\tabcolsep}{5pt}
\renewcommand{\arraystretch}{1.08}

\begin{tabular}{@{}p{1.65cm}p{1.05cm}p{1.7cm}p{10.6cm}@{}}
\toprule
Region & Turn & Speaker & Utterance \\
\midrule

\multicolumn{4}{@{}l}{
\textbf{Case 1: Dialogue 1264; Ground-truth emotion: fear}
} \\

Outside-window clue
& $u_{t-6}$
& Chandler
& \textit{You kissed my best Ross! ... Or something to that effect.} \\

\cmidrule(lr){1-4}

\multirow{4}{*}{Local window}
& $u_{t-4}$
& Chandler
& Really stupid. \\

& $u_{t-3}$
& Mrs. Bing
& Really stupid. \\

& $u_{t-2}$
& Mrs. Bing
& And I don't even know how it happened. \\

& $u_{t-1}$
& Mrs. Bing
& I'm sorry, honey, I promise it will never happen again. \\

Target
& $u_t$
& Mrs. Bing
& \textbf{Are we okay now?} \\

\multicolumn{4}{@{}l}{
Gold: \textbf{fear}
\quad
LoRA: neutral
\quad
Window-local: neutral
\quad
Full DiaRelay: \textbf{fear}
} \\

\midrule

\multicolumn{4}{@{}l}{
\textbf{Case 2: Dialogue 1293; Ground-truth emotion: sadness}
} \\

Outside-window clue
& $u_{t-12}$
& Joey
& \textit{That part was perfect for me! I can't believe I didn't get it!} \\

\cmidrule(lr){1-4}

\multirow{4}{*}{Local window}
& $u_{t-4}$
& Joey
& Come on Ross, be realistic. If I did write something,
what are the chances I could get those guys to star in it? \\

& $u_{t-3}$
& Joey
& Wait a second, I could star in it! \\

& $u_{t-2}$
& Ross
& Or that. \\

& $u_{t-1}$
& Joey
& I can't write! \\

Target
& $u_t$
& Joey
& \textbf{Y'know, I mean I-I-I'm an actor,
I don't have the discipline that takes, y'know?} \\

\multicolumn{4}{@{}l}{
Gold: \textbf{sadness}
\quad
LoRA: fear
\quad
Window-local: fear
\quad
Full DiaRelay: \textbf{sadness}
} \\

\bottomrule
\end{tabular}
\caption{
Case studies on MELD. For each example, we show one
representative historical clue outside the explicit
four-utterance history window, the complete local window, and the
predictions of different variants. Correct predictions are
shown in bold.
}
\label{tab:case_study}
\end{table*}

\subsection{Comparison with State-of-the-Art Methods}
As shown in Table~\ref{tab:main_results}, DiaRelay with Qwen3-4B outperforms the previous best results among backbones no larger than 6B by 1.80\%/2.09\% W-F1/accuracy on IEMOCAP and 0.32/1.33 on MELD. This demonstrates that the proposed relay mechanism remains
effective with a relatively compact backbone.

With Qwen3-8B, DiaRelay further exceeds the previous best MELD results by 0.43\% W-F1 and 0.54\% accuracy. Despite using only a lightweight internal relay memory and no external retrieved or generated knowledge, DiaRelay establishes a new state-of-the-art result on MELD. On IEMOCAP, DiaRelay achieves competitive W-F1 without any external augmentation (InstructERC) or complex graph modeling (MSG-LLM). Unlike most compared methods, DiaRelay obtains nearly identical W-F1 on IEMOCAP and MELD. We conjecture that the smaller training scale of IEMOCAP may provide insufficient supervision for fully optimizing the newly introduced attention-based memory interactions.

Overall, these results show the effectiveness and generalizability of our DiaRelay in enhancing LLM-based emotional understanding across different datasets and different backbone scales.

\subsection{Case Study}
To qualitatively investigate how dialogue-level relay memory
supports emotion recognition beyond the explicit context
window, we present two representative examples from MELD
in Table~\ref{tab:case_study}. In both cases, LoRA and
Window-local DiaRelay make incorrect predictions, whereas
Full DiaRelay correctly identifies the target emotion.

In Case 1, the earlier accusation establishes unresolved interpersonal tension,
making the apparently neutral question ``Are we okay now?''
an expression of fear. Both LoRA and Window-local DiaRelay
predict neutral, whereas Full DiaRelay correctly identifies
fear.

In Case 2, Joey's earlier audition failure provides the
emotional cause underlying his later self-doubt. Without
this earlier context, both LoRA and Window-local DiaRelay
interpret the target as fear, while Full DiaRelay correctly
predicts sadness. These examples are consistent with
dialogue-level relay preserving useful emotional evidence
after it has left the explicit local context.


\subsection{Ablation Studies}
\label{sec:ablation}

We conduct ablation studies to evaluate the overall
effectiveness of DiaRelay and the contributions of its major
design components. Unless otherwise specified, all component
ablations are conducted on MELD with Qwen3-8B using the same
data split, prompt construction, optimization configuration,
random seed, and fixed evaluation checkpoint.

In addition to the LoRA-only baseline, we consider three types
of controlled variants. First, for each target utterance, \emph{Window-local DiaRelay}
reinitializes the relay memory to zero and sequentially writes,
in chronological order, all historical utterances that precede
the target utterance within the current explicit context window
into the memory. The resulting local state is then read by
DaRMR to predict the target utterance, after which the state is
discarded. Thus, the complete SRMT and DaRMR computations
are retained within each window, while the relay state is not
propagated across consecutive sliding windows. Second,
\emph{w/o Error-Corrective Updating} replaces the residual
write signal
$e_t^m = v_t^m - R_{t-1}k_t^m$
with the complete current value $v_t^m$, while retaining the
memory gates and the dual-axis read.Finally, we remove the query-side correction $\Delta q$
and the output-side correction $\Delta o$ separately.

\begin{table}[t]
    \centering
    \small
    \setlength{\tabcolsep}{4.0pt}
    \renewcommand{\arraystretch}{1.08}
    \begin{tabular}{@{}lccccc@{}}
        \toprule
        Variant
        & $\Delta q$
        & $\Delta o$
        & W-F1
        & M-F1
        & Acc. \\
        \midrule

        \multicolumn{6}{l}{\textit{MELD, Qwen3-8B}} \\

        LoRA-only
        & -- & --
        & 67.01 & 52.20 & 68.74 \\

        Window-local DiaRelay
        & \checkmark & \checkmark
        & 68.95 & 55.75 & 70.42 \\

        w/o Error-Corrective Updating
        & \checkmark & \checkmark
        & 69.37 & 54.67 & 70.70 \\

        w/o Output Correction
        & \checkmark & --
        & 69.10 & 54.18 & 70.65 \\

        w/o Query Correction
        & -- & \checkmark
        & 69.41 & 55.42 & 70.84 \\

        Full DiaRelay
        & \checkmark & \checkmark
        & \textbf{70.06}
        & \textbf{57.04}
        & \textbf{71.15} \\

        \midrule
        \multicolumn{6}{l}{\textit{IEMOCAP, Qwen3-8B}} \\

        LoRA-only
        & -- & --
        & 67.43 & 65.30 & 67.59 \\

        Full DiaRelay
        & \checkmark & \checkmark
        & \textbf{70.01}
        & \textbf{68.74}
        & \textbf{69.93} \\

        %
        %
        %

        \bottomrule
    \end{tabular}
    \caption{Ablation results of DiaRelay. Component-level
    ablations are conducted on MELD with Qwen3-8B, while the
    IEMOCAP results evaluate the overall improvement over the
    LoRA-only baseline. W-F1, M-F1, and Acc. denote weighted
    F1, macro F1, and accuracy, respectively. All values are
    percentages.}
    \label{tab:ablation}
\end{table}

\paragraph{Overall effectiveness.}
As shown in Table~\ref{tab:ablation}, the full DiaRelay
consistently outperforms the LoRA-only baseline on both
datasets. On MELD, DiaRelay improves weighted F1, macro F1,
and accuracy by 3.05\%, 4.84\%, and 2.41\%,
respectively. On IEMOCAP, it produces corresponding gains of
2.58\%, 3.44\%, and 2.34\%. These consistent improvements
demonstrate that augmenting LoRA with a dialogue-dependent
relay memory provides substantial benefits beyond static
low-rank adaptation. The particularly clear gains in macro F1
also indicate that the improvement is not limited to dominant
emotion classes.

\paragraph{Effect of Dialogue-level Relay.}
Window-local DiaRelay improves the LoRA-only baseline by
1.94\% weighted F1, 3.55\% macro F1, and 1.68\% accuracy, showing that memory-based aggregation remains useful
even when restricted to the explicit local context. Notably, this
variant does not reduce to LoRA-only: before predicting the target
utterance, its relay state has already been sequentially updated
using the preceding utterances within the current input window
and is therefore generally nonzero.

Nevertheless, this local state is discarded after each prediction
and cannot preserve evidence once it leaves the explicit context
window. Full DiaRelay further surpasses the window-local variant
by 1.11\% weighted F1, 1.29\% macro F1, and 0.73\% accuracy. Since the two variants use the same explicit context,
SRMT update, and dual-axis memory read, while differing in
whether the relay state is propagated across consecutive windows,
the performance gap demonstrates the benefit of persistent
cross-window memory propagation.

\paragraph{Effect of Error-corrective Updating.}
Replacing the error-corrective updating signal with the complete
current value reduces weighted F1 from 70.06\% to 69.37\%, macro
F1 from 57.04\% to 54.67\%, and accuracy from 71.15\% to 70.70\%.
The degradation is especially pronounced in macro F1, with a
drop of 2.37\%. These results support the residual updating
strategy used in Selective Relay Memory Transition. Rather
than repeatedly accumulating the complete current value,
writing only the component that is not recovered from the
previous memory produces a more effective dialogue state.

\paragraph{Effect of dual-axis memory read.}
Removing either correction path consistently degrades all three
metrics, confirming that both $\Delta q$ and $\Delta o$ contribute
to the final prediction. The output-only variant outperforms the
query-only variant by 0.31\%, 1.24\%, and 0.19\% in weighted F1,
macro F1, and accuracy, respectively, indicating a stronger
standalone contribution from the output-side path. Nevertheless,
the full dual-axis read improves over the output-only variant by
0.65\%, 1.62\%, and 0.31\%, and over the query-only variant by
0.96\%, 2.86\%, and 0.50\%. These results demonstrate that the two
correction paths provide complementary historical guidance,
particularly for class-balanced macro F1.

\textbf{Due to space limitations, further mechanism analysis of DiaRelay and the ablation study of relay rank in DiaRelay are provided in Supplementary Material B1, B2}.

\section{Limitations and Future Work}
Although DiaRelay achieves strong performance on MELD and IEMOCAP, we evaluate it primarily under the text-only ERC setting. Its applicability to multimodal conversational understanding and more general dialogue tasks therefore remains to be explored. In addition, the current framework focuses on classification performance and does not explicitly identify the historical evidence associated with each prediction or provide a natural-language explanation for the predicted emotion. Future work will extend DiaRelay to multimodal and general-purpose dialogue settings, and incorporate an explanation generation mechanism that jointly produces emotion predictions and faithful, human-readable rationales grounded in the relayed dialogue memory.

\section{Conclusion}

In this paper, we present DiaRelay, a lightweight memory-augmented plug-in that equips large language models with persistent dialogue-level memory without extending the explicit context window. By maintaining a constant-size relay state across successive utterances, DiaRelay allows useful historical emotional evidence to continue influencing later predictions after it leaves the local input window. Extensive experiments on MELD and IEMOCAP with Qwen3-4B and Qwen3-8B show the effectiveness and generalizability of DiaRelay for LLM-based emotion recognition.
Moreover, DiaRelay retains the original autoregressive training objective and standard inference pipeline of the backbone, without requiring auxiliary memory supervision, external retrieval, generated knowledge, speaker profiles, or test-time parameter updates. Overall, DiaRelay provides a new lightweight solution and establishes
a strong history-only baseline for emotion recognition in conversation.

\bibliography{aaai2027}

@article{erc_survey,
  author  = {Poria, Soujanya and Majumder, Navonil and Mihalcea, Rada and Hovy, Eduard},
  title   = {Emotion Recognition in Conversation: Research Challenges, Datasets, and Recent Advances},
  journal = {IEEE Access},
  volume  = {7},
  pages   = {100943--100953},
  year    = {2019},
  doi     = {10.1109/ACCESS.2019.2929050}
}

@inproceedings{dialoguernn,
  author    = {Majumder, Navonil and Poria, Soujanya and Hazarika, Devamanyu and Mihalcea, Rada and Gelbukh, Alexander and Cambria, Erik},
  title     = {{DialogueRNN}: An Attentive {RNN} for Emotion Detection in Conversations},
  booktitle = {Proceedings of the AAAI Conference on Artificial Intelligence},
  volume    = {33},
  number    = {1},
  pages     = {6818--6825},
  year      = {2019},
  doi       = {10.1609/aaai.v33i01.33016818}
}

@inproceedings{dialoguegcn,
  author    = {Ghosal, Deepanway and Majumder, Navonil and Poria, Soujanya and Chhaya, Niyati and Gelbukh, Alexander},
  title     = {{DialogueGCN}: A Graph Convolutional Neural Network for Emotion Recognition in Conversation},
  booktitle = {Proceedings of the 2019 Conference on Empirical Methods in Natural Language Processing and the 9th International Joint Conference on Natural Language Processing},
  pages     = {154--164},
  year      = {2019},
  publisher = {Association for Computational Linguistics},
  doi       = {10.18653/v1/D19-1015}
}

@inproceedings{dialoguecrn,
  author    = {Hu, Dou and Wei, Lingwei and Huai, Xiaoyong},
  title     = {{DialogueCRN}: Contextual Reasoning Networks for Emotion Recognition in Conversations},
  booktitle = {Proceedings of the 59th Annual Meeting of the Association for Computational Linguistics and the 11th International Joint Conference on Natural Language Processing},
  pages     = {7042--7052},
  year      = {2021},
  publisher = {Association for Computational Linguistics},
  doi       = {10.18653/v1/2021.acl-long.547}
}

@inproceedings{dagerc,
  author    = {Shen, Weizhou and Wu, Siyue and Yang, Yunyi and Quan, Xiaojun},
  title     = {Directed Acyclic Graph Network for Conversational Emotion Recognition},
  booktitle = {Proceedings of the 59th Annual Meeting of the Association for Computational Linguistics and the 11th International Joint Conference on Natural Language Processing},
  pages     = {1551--1560},
  year      = {2021},
  publisher = {Association for Computational Linguistics},
  doi       = {10.18653/v1/2021.acl-long.123}
}

@inproceedings{dialogxl,
  author    = {Shen, Weizhou and Chen, Junqing and Quan, Xiaojun and Xie, Zhixian},
  title     = {{DialogXL}: All-in-One {XLNet} for Multi-Party Conversation Emotion Recognition},
  booktitle = {Proceedings of the AAAI Conference on Artificial Intelligence},
  volume    = {35},
  number    = {15},
  pages     = {13789--13797},
  year      = {2021},
  doi       = {10.1609/aaai.v35i15.17625}
}

@inproceedings{compm,
  author    = {Lee, Joosung and Lee, Wooin},
  title     = {{CoMPM}: Context Modeling with Speaker's Pre-trained Memory Tracking for Emotion Recognition in Conversation},
  booktitle = {Proceedings of the 2022 Conference of the North American Chapter of the Association for Computational Linguistics: Human Language Technologies},
  pages     = {5669--5679},
  year      = {2022},
  publisher = {Association for Computational Linguistics},
  doi       = {10.18653/v1/2022.naacl-main.416}
}

@article{instructerc,
  author  = {Lei, Shanglin and Dong, Guanting and Wang, Xiaoping and Wang, Keheng and Qiao, Runqi and Wang, Sirui},
  title   = {{InstructERC}: Reforming Emotion Recognition in Conversation with Multi-task Retrieval-Augmented Large Language Models},
  journal = {arXiv preprint arXiv:2309.11911},
  year    = {2023}
}

@inproceedings{laercs,
  author    = {Fu, Yumeng and Wu, Junjie and Wang, Zhongjie and Zhang, Meishan and Shan, Lili and Wu, Yulin and Liu, Bingquan},
  title     = {{LaERC-S}: Improving {LLM}-based Emotion Recognition in Conversation with Speaker Characteristics},
  booktitle = {Proceedings of the 31st International Conference on Computational Linguistics},
  pages     = {6748--6761},
  year      = {2025},
  publisher = {Association for Computational Linguistics}
}

@inproceedings{coe,
  author    = {Shen, Zhiyu and Pang, Yunhe and Rao, Yanghui and Yu, Jianxing},
  title     = {{CoE}: A Clue of Emotion Framework for Emotion Recognition in Conversations},
  booktitle = {Proceedings of the 63rd Annual Meeting of the Association for Computational Linguistics},
  pages     = {23548--23563},
  year      = {2025},
  publisher = {Association for Computational Linguistics},
  doi       = {10.18653/v1/2025.acl-long.1148}
}

@inproceedings{prcemo,
  author    = {Li, Xinran and Liu, Yu and Qiao, Jiaqi and Xu, Xiujuan},
  title     = {Do {LLM}s Feel? Teaching Emotion Recognition with Prompts, Retrieval, and Curriculum Learning},
  booktitle = {Proceedings of the AAAI Conference on Artificial Intelligence},
  volume    = {40},
  number    = {38},
  pages     = {31778--31786},
  year      = {2026},
  doi       = {10.1609/aaai.v40i38.40446}
}

@inproceedings{lora,
  author    = {Hu, Edward J. and Shen, Yelong and Wallis, Phillip and Allen-Zhu, Zeyuan and Li, Yuanzhi and Wang, Shean and Wang, Lu and Chen, Weizhu},
  title     = {{LoRA}: Low-Rank Adaptation of Large Language Models},
  booktitle = {International Conference on Learning Representations},
  year      = {2022}
}

@article{busso2008iemocap,
  author  = {Busso, Carlos and Bulut, Murtaza and Lee, Chi-Chun and Kazemzadeh, Abe and Mower, Emily and Kim, Samuel and Chang, Jeannette N. and Lee, Sungbok and Narayanan, Shrikanth S.},
  title   = {{IEMOCAP}: Interactive Emotional Dyadic Motion Capture Database},
  journal = {Language Resources and Evaluation},
  volume  = {42},
  number  = {4},
  pages   = {335--359},
  year    = {2008},
  doi     = {10.1007/s10579-008-9076-6}
}

@inproceedings{poria2019meld,
  author    = {Poria, Soujanya and Hazarika, Devamanyu and Majumder, Navonil and Naik, Gautam and Cambria, Erik and Mihalcea, Rada},
  title     = {{MELD}: A Multimodal Multi-Party Dataset for Emotion Recognition in Conversations},
  booktitle = {Proceedings of the 57th Annual Meeting of the Association for Computational Linguistics},
  pages     = {527--536},
  year      = {2019},
  publisher = {Association for Computational Linguistics},
  doi       = {10.18653/v1/P19-1050}
}

@inproceedings{cosmic,
  author    = {Ghosal, Deepanway and Majumder, Navonil and Gelbukh, Alexander and Mihalcea, Rada and Poria, Soujanya},
  title     = {{COSMIC}: {CO}mmon{S}ense Knowledge for e{M}otion Identification in Conversations},
  booktitle = {Findings of the Association for Computational Linguistics: EMNLP 2020},
  pages     = {2470--2481},
  year      = {2020},
  publisher = {Association for Computational Linguistics},
  doi       = {10.18653/v1/2020.findings-emnlp.224}
}

@article{mmla,
  author  = {Zhang, Hanlei and Li, Zhuohang and Zhu, Yeshuang and Xu, Hua and Wang, Peiwu and Zhang, Jinchao and Zhou, Jie and Zhu, Haige},
  title   = {Can Large Language Models Help Multimodal Language Analysis? {MMLA}: A Comprehensive Benchmark},
  journal = {arXiv preprint arXiv:2504.16427},
  year    = {2025}
}

@inproceedings{mseadapter,
  author    = {Yang, Yang and Dong, Xunde and Qiang, Yupeng},
  title     = {{MSE-Adapter}: A Lightweight Plugin Endowing {LLM}s with the Capability to Perform Multimodal Sentiment Analysis and Emotion Recognition},
  booktitle = {Proceedings of the AAAI Conference on Artificial Intelligence},
  volume    = {39},
  number    = {24},
  pages     = {25642--25650},
  year      = {2025},
  doi       = {10.1609/aaai.v39i24.34755}
}

@inproceedings{bioserc,
  author    = {Xue, Jieying and Nguyen, Minh-Phuong and Matheny, Blake and Nguyen, Le-Minh},
  title     = {{BiosERC}: Integrating Biography Speakers Supported by {LLM}s for {ERC} Tasks},
  booktitle = {Artificial Neural Networks and Machine Learning -- ICANN 2024},
  pages     = {277--292},
  year      = {2024},
  publisher = {Springer Nature Switzerland}
}

@article{speechcuellm,
  author  = {Wu, Zehui and Gong, Ziwei and Ai, Lin and Shi, Pengyuan and Donbekci, Kaan and Hirschberg, Julia},
  title   = {Beyond Silent Letters: Amplifying {LLM}s in Emotion Recognition with Vocal Nuances},
  journal = {arXiv preprint arXiv:2407.21315},
  year    = {2024}
}

@inproceedings{causalerc,
  author    = {Jing, Ran and Tu, Geng and Zhang, Yice and Xu, Ruifeng},
  title     = {{Causal-ERC}: A Multimodal Framework with Causal Prompting for Emotion Recognition in Conversations with Large Language Models},
  booktitle = {Proceedings of the AAAI Conference on Artificial Intelligence},
  volume    = {40},
  number    = {37},
  pages     = {31383--31391},
  year      = {2026},
  doi       = {10.1609/aaai.v40i37.40402}
}

@inproceedings{ding-etal-2025-msg,
    title = "{MSG}-{LLM}: A Multi-scale Interactive Framework for Graph-enhanced Large Language Models",
    author = "Ding, Jiayu  and
      Zheng, Zhangkai  and
      Lin, Benshuo  and
      Xue, Yun  and
      Song, Yiping",
    booktitle = "Proceedings of the 31st International Conference on Computational Linguistics",
    month = jan,
    year = "2025",
    address = "Abu Dhabi, UAE",
    publisher = "Association for Computational Linguistics",
    url = "https://aclanthology.org/2025.coling-main.648/",
    pages = "9687--9700"
}

@article{qwen3,
  title   = {Qwen3 Technical Report},
  author  = {Yang, An and Li, Anfeng and Yang, Baosong and others},
  journal = {arXiv preprint arXiv:2505.09388},
  year    = {2025}
}

@inproceedings{schlag2021linear,
  title     = {Linear Transformers Are Secretly Fast Weight Programmers},
  author    = {Schlag, Imanol and Irie, Kazuki and Schmidhuber, J{\"u}rgen},
  booktitle = {Proceedings of the 38th International Conference on Machine Learning},
  pages     = {9355--9366},
  year      = {2021}
}

\clearpage
\appendix
\newcommand{\suppsection}[2]{\section*{#1. #2}}
\renewcommand{\thetable}{S\arabic{table}}
\renewcommand{\thefigure}{S\arabic{figure}}
\renewcommand{\theequation}{S\arabic{equation}}

This supplementary material provides additional experimental
details and analyses omitted from the main paper due to the page
limit. Section A1 briefly introduces the compared baselines, and
Section A2 provides further implementation details. Section B1
presents a mechanism analysis and layer-wise diagnostics of
DiaRelay, while Section B2 studies the effect of the relay rank.

\suppsection{A1}{Brief Introduction to Baselines}

Table~\ref{tab:supp-baseline-settings} summarizes the contextual
information, auxiliary resources, modalities, and result sources
of the methods compared in the main paper. ``Future/full
dialogue'' indicates that the reported setting can use utterances
after the current target or representations constructed from the
complete dialogue. ``External knowledge'' includes retrieved
demonstrations, generated speaker descriptions, commonsense
knowledge, biographies, or other speaker-related information
beyond the original dialogue text. The indicators are consistent
with those used in Table~2 of the main paper. The brief introduction of all baselines is as follows:

\noindent\textbf{DialogueRNN and DialogueGCN:}
DialogueRNN models conversational dynamics with recurrent
speaker-aware states, while DialogueGCN represents utterances as
graph nodes and propagates information through conversational
relations. Their reported results are obtained under settings that
use full-dialogue contextual information.

\noindent\textbf{COSMIC:}
COSMIC augments contextual emotion modeling with commonsense
knowledge. We therefore mark it as using both full-dialogue
information and additional external knowledge.

\noindent\textbf{MMLA:}
MMLA is included as a recent LLM-based benchmark. We report its
supervised fine-tuning result with the Llama-3.2-3B backbone under
the setting reported by the original work.

\noindent\textbf{MSE-Adapter:}
MSE-Adapter is a lightweight multimodal adapter evaluated with
textual, acoustic, and visual information. In the comparison
table, it is marked as using future/full-dialogue information
according to the input construction used for the reported result.

\noindent\textbf{InstructERC:}
InstructERC reformulates ERC as an instruction-following task and
incorporates retrieved demonstrations together with auxiliary
speaker- and emotion-related supervision. Its result is therefore
marked as using additional external information.

\noindent\textbf{BiosERC-7B:}
BiosERC-7B introduces speaker biographies generated or collected
beyond the original target dialogue. The reported setting also
uses full-dialogue information and is marked accordingly.

\noindent\textbf{MSG-LLM:}
MSG-LLM combines large language models with multi-scale
graph-enhanced conversational modeling and additionally
incorporates speaker-related information. It is therefore marked
as using external knowledge.

\noindent\textbf{LaERC-S:}
LaERC-S introduces speaker characteristics, such as mental states
and behavioral descriptions, through a multi-stage learning
procedure. Its reported setting uses both full-dialogue
information and additional speaker-related knowledge.

\noindent\textbf{SpeechCueLLM:}
SpeechCueLLM incorporates vocal cues in addition to dialogue
text. We include it as a text--audio comparison method and report
the result from the original work.

\noindent\textbf{Causal-ERC (T):}
Causal-ERC is designed for causal conversational emotion
recognition. We include its text-only result to provide a
comparison under a causal information setting without additional
audio or visual inputs.

\noindent\textbf{PRC-Emo (Causal):}
The original PRC-Emo framework combines emotion-sensitive
prompting, demonstration retrieval, and curriculum learning. For
a fair history-only comparison, we construct a causal version by
excluding all future utterances from the context of the current
target while retaining the remaining prompting and training
procedures. Qwen2.5-7B is used for IEMOCAP and Qwen3-8B is used
for MELD, following the backbone assignment reported in the main
paper. All PRC-Emo (Causal) scores in Table~2 of the main paper
are obtained from this reimplementation.

\begin{table*}[t]
\centering
\caption{Information settings of the compared methods. T, A, and
V denote text, audio, and visual modalities, respectively.}
\label{tab:supp-baseline-settings}
\small
\setlength{\tabcolsep}{7pt}
\renewcommand{\arraystretch}{1.03}
\begin{tabular}{@{}lcccl@{}}
\toprule
Method & Future/full dialogue & External knowledge
       & Modality & Result source \\
\midrule
DialogueRNN      & Yes & No  & T   & Published result \\
DialogueGCN      & Yes & No  & T   & Published result \\
COSMIC           & Yes & Yes & T   & Published result \\
MMLA (SFT)       & No  & No  & T   & Published result \\
MSE-Adapter      & Yes & No  & TAV & Published result \\
InstructERC      & No  & Yes & T   & Published result \\
BiosERC-7B       & Yes & Yes & T   & Published result \\
MSG-LLM          & No  & Yes & T   & Published result \\
LaERC-S          & Yes & Yes & T   & Published result \\
SpeechCueLLM     & No  & No  & TA  & Published result \\
Causal-ERC (T)   & No  & No  & T   & Published result \\
PRC-Emo (Causal) & No  & Yes & T   & Causal reimplementation \\
\bottomrule
\end{tabular}
\end{table*}

\suppsection{A2}{More Implementation Details}

\subsection*{Dataset Preprocessing and Label Space}

We use the official dialogue-level splits of IEMOCAP and MELD.
The training and validation portions are combined for model
fitting, while the official test split is kept unchanged for
evaluation and reporting. Each dialogue is processed in its
original chronological order. The relay memories of all adapted
layers are initialized to zero at the beginning of each dialogue
and reset before processing the next dialogue.

For the prediction of the $t$-th utterance, the explicit input
contains the current target utterance and at most four immediately
preceding utterances, resulting in a maximum window length of
five utterances. No future utterance is included. MELD retains
the original speaker names provided by the dataset. For IEMOCAP,
the original session and gender identifiers are converted to
fixed session-specific speaker names so that the two speakers
remain distinguishable throughout each dialogue.

The exact label strings used during both training and evaluation
are: \texttt{neutral}, \texttt{surprise}, \texttt{fear},
\texttt{sadness}, \texttt{joy}, \texttt{disgust}, and
\texttt{anger} for \textbf{MELD}, and \texttt{happy}, \texttt{sad}, \texttt{neutral}, \texttt{angry},
\texttt{excited}, and \texttt{frustrated} for \textbf{IEMOCAP}.

At inference time, the decoded text following the assistant
marker and preceding the end-of-message token is stripped of
surrounding whitespace. A prediction is treated as valid only
when it exactly matches one of the dataset-specific label strings
above. We do not apply lowercase conversion, semantic alias
mapping, or post-hoc correction to unmatched outputs. Unmatched
generations are counted as incorrect predictions.

\subsection*{Prompt Construction}

For each target utterance, we construct a chat-formatted input
using the following template. The target utterance is included in
the chronological context and repeated in the user query.
Candidate labels are not explicitly enumerated in the prompt.
The prompt template we used is as below:

\begingroup
\small
\raggedright
\setlength{\parindent}{0pt}
\textbf{System:}\\
\texttt{\#\#\# You are an expert at analyzing the}\\
\texttt{emotion of utterances among speakers in}\\
\texttt{a conversation.}\\[2pt]
\texttt{\#\#\# Given the following conversation as}\\
\texttt{a context}\\
\texttt{[Speaker$_{t-4}$]: [utterance$_{t-4}$]}\\
\texttt{$\cdots$}\\
\texttt{[Speaker$_{t-1}$]: [utterance$_{t-1}$]}\\
\texttt{[Speaker$_t$]: [target utterance$_t$]}\\[2pt]
\textbf{User:}\\
\texttt{Based on above conversation, which}\\
\texttt{emotional label of [Speaker$_t$] in the}\\
\texttt{utterance ``[target utterance$_t$]''.}\\[2pt]
\textbf{Assistant during training:}\\
\texttt{[ground-truth label]}
\endgroup

The same prompt construction and explicit context window are
used by LoRA-only, Window-local DiaRelay, the
component-ablation variants, and Full DiaRelay. Although the
target is repeated in the user query, DiaRelay identifies its
token span internally and applies the memory-conditioned
corrections only to the target-utterance token positions.

\subsection*{Model and Optimization Configuration}

Table~\ref{tab:supp-configuration} summarizes the shared
implementation settings. The backbone is loaded with 4-bit NF4
quantization and bfloat16 computation. LoRA is applied to all
linear modules, while DiaRelay is inserted into all Transformer
layers. Each adapted layer maintains an independent $r\times r$
relay memory. This memory is a dialogue-dependent activation rather
than a trainable parameter, and its size remains constant as the
dialogue grows.

For Qwen3-8B, the backbone contains 8,190,735,360 parameters.
LoRA introduces 87,293,952 trainable parameters, while DiaRelay only
adds an extra 7,078,176 trainable parameters. The resulting model contains
94,372,128 trainable parameters in total, while the backbone
remains frozen. Therefore, the additional DiaRelay parameters
account for approximately 0.09\% of the Qwen3-8B backbone.

\begin{table*}[t]
\centering
\caption{Model, optimization, and decoding configuration used in
the main DiaRelay experiments.}
\label{tab:supp-configuration}
\small
\setlength{\tabcolsep}{6pt}
\renewcommand{\arraystretch}{1.02}
\begin{tabular}{@{}p{0.29\textwidth}p{0.62\textwidth}@{}}
\toprule
Item & Setting \\
\midrule
Backbones
& Qwen3-4B and Qwen3-8B \\
Input modality
& Text only \\
Explicit context
& At most four preceding utterances plus the current target \\
Backbone quantization
& 4-bit NF4 with double quantization \\
Computation precision
& bfloat16; relay memories stored in FP32 \\
LoRA rank / alpha / dropout
& 32 / 128 / 0.05 \\
LoRA target modules
& All linear modules; no trainable bias \\
Relay rank $r$
& 8 in the main experiments \\
Relay scaling factor
& 16 \\
Adapted layers
& All Transformer layers \\
Memory initialization
& Zero at each dialogue boundary \\
Optimizer
& AdamW \\
Per-device micro-batch
& One target utterance \\
LoRA learning rate
& $3\times10^{-4}$ \\
DiaRelay learning rate
& $2\times10^{-4}$ \\
Weight decay
& 0 \\
Learning-rate schedules
& Linear for LoRA; cosine for DiaRelay \\
Warmup ratios
& 0.03 for LoRA; 0.10 for DiaRelay \\
Gradient accumulation
& Four target utterances per optimizer update \\
Maximum sequence length
& 2048 tokens \\
Gradient clipping
& 0.3 \\
Attention implementation
& SDPA \\
Gradient checkpointing
& Disabled for dialogue-stream training \\
Decoding
& Greedy decoding without sampling \\
Maximum generated tokens
& 10 \\
Learning objective
& Autoregressive negative log-likelihood \\
Test-time parameter updates
& None \\
\bottomrule
\end{tabular}
\end{table*}

\subsection*{Training and Evaluation Protocol}

All experiments are conducted on a single NVIDIA RTX 3090 GPU. Following PRC-Emo, the train and validation splits are merged, and the official test split is evaluated at the end of each epoch. We use fixed reporting checkpoints rather than selecting the numerically best
test result. For both Qwen3-8B and Qwen3-4B on MELD, we report
the third-epoch checkpoint, whereas for both backbones on
IEMOCAP, we report the eighth-epoch checkpoint. Because the two
datasets contain different numbers of training instances per
epoch, these dataset-specific epoch indices correspond to the
same predefined optimization-step budget.

The main results follow the multi-run evaluation protocol
described in the main paper. To limit the computational cost of
fine-tuning large language models, the auxiliary controlled runs,
including the LoRA-only reference, component-ablation variants,
and alternative relay-rank settings, are conducted with a fixed
random seed. For these runs, the data split, prompt construction,
explicit context window, optimization configuration, and
reporting rule are kept unchanged.

\subsection*{Sequential Processing Procedure}

DiaRelay follows a read-before-write procedure. For a dialogue
containing $T$ utterances,
$D=\{(x_t,s_t)\mid t=1,\ldots,T\}$, where $x_t$ and $s_t$
denote the $t$-th utterance and its speaker identity,
respectively, training and inference proceed as follows:

\begin{enumerate}
    \item Initialize $R^{(\ell)}_0=0$ for every adapted layer
    $\ell$.

    \item Construct the explicit context $\mathcal{C}_t$ from the
    current target utterance and at most four preceding
    utterances.

    \item At each adapted layer, DaRMR reads only the previous
    relay memory $R^{(\ell)}_{t-1}$ and produces query-side and
    output-side corrections for the target span.

    \item The backbone predicts the emotion label of the current
    target utterance.

    \item The target-span representation $u^{(\ell)}_t$ is mean
    pooled and projected into the relay space.

    \item SRMT updates the memory from $R^{(\ell)}_{t-1}$ to
    $R^{(\ell)}_t$ using the gated error-corrective transition.

    \item The updated memory is passed to the prediction of the
    next utterance. After the dialogue ends, all relay memories are
    reset to zero.
\end{enumerate}

This order ensures that the current utterance cannot be written
into the relay memory before its own prediction. It can only
affect subsequent utterances, preserving the causal history-only
setting.

\paragraph{Cross-utterance gradient span.}
During dialogue-stream training, the relay memory is propagated
throughout the entire dialogue. However, retaining the complete
cross-utterance computation graph would result in
substantial memory consumption. Thus, we employ truncated
backpropagation across utterance transitions. Let $K$ denote
the number of consecutive utterance computations whose relay
memories remain connected in the same autograd graph before the
memory is detached. Importantly, $K$ controls only the
cross-utterance gradient span and does not limit the amount of
dialogue history carried by the relay memory during forward
propagation or inference.

With $K=1$, the relay memory is detached after every utterance,
and no gradient is propagated through an inter-utterance memory
transition. With $K=2$, the prediction loss of the subsequent
utterance can directly optimize the preceding relay-memory
update. For $K>1$, we additionally apply the offset segmentation
protocol to reduce the dependence on a single set of fixed
segment boundaries.

\begin{table}[t]
\centering
\caption{Effect of the cross-utterance gradient span $K$ on
MELD with Qwen3-8B. All results are reported at the fixed
third epoch. $K$ controls the differentiable span during
training rather than the forward memory horizon.}
\label{tab:gradient_span}
\small
\setlength{\tabcolsep}{8pt}
\renewcommand{\arraystretch}{1.08}
\begin{tabular}{@{}cccc@{}}
\toprule
Gradient span $K$ & W-F1 & M-F1 & Acc. \\
\midrule
1 & 68.82 & 54.01 & 69.50 \\
\textbf{2} & \textbf{70.06} & \textbf{57.04}
            & \textbf{71.15} \\
3 & 68.30 & 54.32 & 69.69 \\
4 & 69.07 & 54.75 & 70.61 \\
\bottomrule
\end{tabular}
\end{table}

As shown in Table~\ref{tab:gradient_span}, setting $K=2$
achieves the strongest overall performance. Compared with
$K=1$, it improves weighted F1, macro F1, and accuracy by
1.24, 3.03, and 1.65 points, respectively. This result
indicates that enabling gradients to pass through an adjacent
relay transition is beneficial for jointly optimizing memory
writing and subsequent memory utilization.

Further extending the differentiable span to $K=3$ or $K=4$
does not provide additional improvements. In particular,
$K=2$ exceeds $K=3$ by 1.76 weighted-F1 points and exceeds
$K=4$ by 0.99 points. These results suggest that a longer
cross-utterance gradient chain is not necessarily more
effective. Therefore, we therefore use $K=2$, which provides effective
cross-utterance credit assignment while keeping the training
graph compact.

\suppsection{B1}{Mechanism Analysis of DiaRelay}

Here, we analyze how the two memory-readout paths in DaRMR affect
the self-attention computation of the backbone. The query-side
and output-side paths intervene at different stages of
self-attention and therefore provide complementary forms of
historical guidance. For clarity, we omit the layer superscript
$(\ell)$ in the following derivation when no ambiguity arises.

For a target-token position $i$, let
\begin{equation}
q^{\mathrm{base}}_{t,i}
=
q^0_{t,i}+\Delta q^L_{t,i}
\end{equation}
denote the query produced by the frozen backbone and its LoRA
correction. After introducing the query-side memory correction,
the query becomes
\begin{equation}
\widetilde{q}_{t,i}
=
q^{\mathrm{base}}_{t,i}+\Delta q^m_t.
\end{equation}

For a visible token position $j$ with key $k_{t,j}$, the original
attention logit is
\begin{equation}
a^{\mathrm{base}}_{i,j}
=
\frac{
\left(q^{\mathrm{base}}_{t,i}\right)^{\top}k_{t,j}
}{
\sqrt{d_h}
},
\end{equation}
whereas the memory-conditioned logit can be written as
\begin{equation}
a^{\mathrm{relay}}_{i,j}
=
a^{\mathrm{base}}_{i,j}
+
\underbrace{
\frac{
\left(\Delta q^m_t\right)^{\top}k_{t,j}
}{
\sqrt{d_h}
}
}_{\delta_{i,j}}.
\end{equation}

Accordingly, the new attention probability satisfies
\begin{equation}
p^{\mathrm{relay}}_{i,j}
=
\frac{
p^{\mathrm{base}}_{i,j}\exp(\delta_{i,j})
}{
\sum_{j'}
p^{\mathrm{base}}_{i,j'}\exp(\delta_{i,j'})
}.
\end{equation}

This expression shows that the query-side path does not append
new visible tokens or directly replace the pretrained attention
pattern. Instead, it introduces a history-conditioned logit bias
whose effect depends on the alignment between the retrieved
memory correction and each visible key. Tokens aligned with the
relayed historical evidence receive relatively larger attention
weights, while inconsistent tokens are relatively downweighted.
Because the correction is applied only to target-token queries,
the representations of the explicit historical context remain
unchanged.

\begin{figure*}[t]
    \centering
    \includegraphics[width=0.47\textwidth]
    {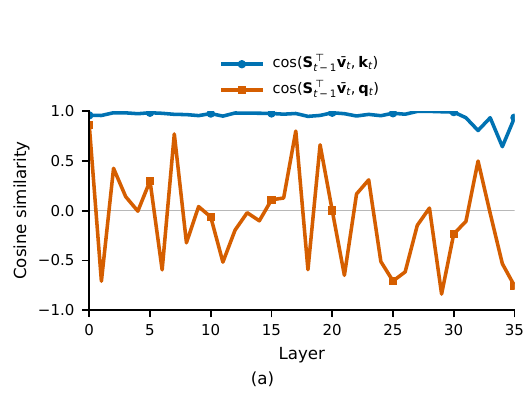}
    \hfill
    \includegraphics[width=0.47\textwidth]
    {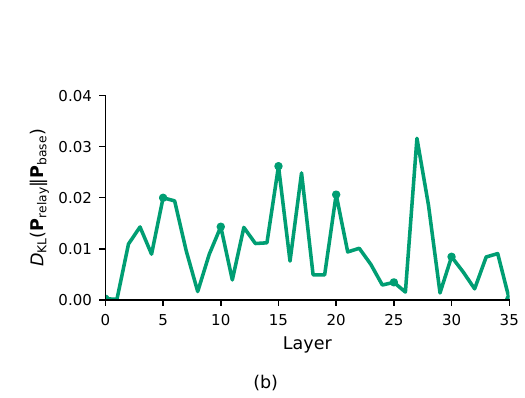}
    \caption{Layer-wise diagnostics of the query-side memory
    read in DaRMR. (a) Cosine similarity between the
    memory-derived address and the current key and query
    representations. The symbol $S_{t-1}$ shown in the panel
    denotes the same relay memory written as $R_{t-1}$ in the
    paper. (b) KL divergence between the relay-conditioned and
    base attention distributions across Transformer layers. The
    panels are used as mechanism diagnostics rather than
    standalone performance comparisons.}
    \label{fig:supp-query-diagnostics}
\end{figure*}

Figure~\ref{fig:supp-query-diagnostics} provides a direct
diagnostic view of the query-side memory read. In
Figure~\ref{fig:supp-query-diagnostics}(a), the memory-derived
direction $R_{t-1}^{\top}\bar{v}_t$ remains strongly aligned with
the current key representation, whereas its similarity with the
query representation varies substantially across layers. This
pattern is consistent with the value-conditioned addressing used
in DaRMR: the relay memory transforms the current value
representation into a direction that can interact with the key
space when injected into the target query.

Figure~\ref{fig:supp-query-diagnostics}(b) further shows that the
query-side correction produces nonzero but generally small
changes in the attention distribution, with larger deviations
concentrated in a subset of layers. This layer-dependent pattern
supports the interpretation given by Eq.~(S5). The correction
does not reconstruct or replace the backbone attention
distribution; instead, it selectively adjusts the relative
weighting of currently visible evidence. The query-side branch
therefore acts as a lightweight routing signal conditioned on
the relayed history.

The output-side path operates at a different stage. Let $o^{\mathrm{base}}_{t,i}$ denote the original attention output and its LoRA residual. It is obtained as follows:
\begin{equation}
o^{\mathrm{base}}_{t,i}
=
o^0_{t,i}+\Delta o^L_{t,i}
\end{equation}
DaRMR adds the memory-conditioned output correction after
attention aggregation:
\begin{equation}
\begin{aligned}
\widetilde{o}_{t,i} &= o^{\mathrm{base}}_{t,i}+\Delta o^m_t, \\
\Delta o^m_t &= \frac{\alpha_o}{r}P_oR_{t-1}q^m_t.
\end{aligned}
\label{eq:o}
\end{equation}
According to Equation (\ref{eq:o}), the output-side correction directly supplements the aggregated
target representation with information retrieved from the
historical relay memory, rather than modifying the
attention probabilities. In other words, the query-side path
controls how currently visible evidence is weighted, whereas the
output-side path adds a history-conditioned residual after the
visible evidence has been aggregated.

Thus, the two paths are complementary rather than redundant. The
component ablations in Table~4 of the main paper support this
insight: removing either path degrades all three metrics,
while the complete dual-axis read achieves the strongest overall
performance. The output-only variant performs better than the
query-only variant when each path is used separately, suggesting
that direct historical supplementation has a stronger standalone
contribution. Nevertheless, combining it with query-side
reweighting produces further gains, especially in macro F1,
showing that historical retrieval and local-evidence routing work
together in DaRMR.

\suppsection{B2}{Ablation Study of Relay Rank}

The relay rank $r$ controls both the dimensionality of the relay
vectors and the size of the per-layer memory
$R_t\in\mathbb{R}^{r\times r}$. To study the
capacity--compactness trade-off, we evaluate
$r\in\{4,8,16,32\}$ on MELD with Qwen3-8B. All configurations
use the same data split, prompt construction, explicit context
window, training schedule, and reporting rule. Only the relay
rank is changed. Following the MELD reporting protocol, all
variants are evaluated at the third-epoch checkpoint. The main
experiments use $r=8$.

\begin{table}[t]
\centering
\caption{Effect of the relay rank on MELD with Qwen3-8B. Memory
entries denote the number of elements in the $r\times r$ relay
memory maintained by each adapted layer.}
\label{tab:supp-relay-rank}
\small
\setlength{\tabcolsep}{5pt}
\renewcommand{\arraystretch}{1.02}
\begin{tabular}{@{}ccccc@{}}
\toprule
Rank $r$ & Memory entries & W-F1 & M-F1 & Acc. \\
\midrule
4  & 16   & 68.47 & 54.34 & 70.27 \\
8  & 64   & \textbf{70.06} & \textbf{57.04}
          & \textbf{71.15} \\
16 & 256  & 69.63 & 55.46 & 71.00 \\
32 & 1024 & 69.84 & 55.96 & 70.84 \\
\bottomrule
\end{tabular}
\end{table}

As shown in Table~\ref{tab:supp-relay-rank}, increasing the relay
rank from 4 to 8 improves weighted F1, macro F1, and accuracy by
1.59, 2.70, and 0.88 points, respectively. This indicates that an
excessively small relay space limits the capacity to preserve
diverse dialogue evidence. However, further increasing the rank
to 16 or 32 does not produce additional gains. The $r=8$ setting
exceeds $r=16$ by 0.43/1.58/0.15 points and exceeds $r=32$ by
0.22/1.08/0.31 points in weighted F1, macro F1, and accuracy,
respectively. These results show that simply enlarging the relay
memory is not sufficient to improve performance and that $r=8$
provides the strongest overall result with a compact per-layer
memory. Therefore, we use $r=8$ as the default setting in the main
experiments.


\end{document}